\documentclass{esannV2}
\usepackage[utf8]{inputenc}
\usepackage{amssymb,amsmath,array}
\usepackage{booktabs}
\usepackage{multirow}
\usepackage{graphicx}

\begin{document}

\title{TimePred: efficient and interpretable offline change point detection for high volume data - with application to industrial process monitoring}

\author{Simon Leszek
%
\thanks{This work was funded by the German Federal Ministry for Research, Technology and Space through the project RIESIQ (under ref 16IS24087A).}
%
\vspace{.3cm}\\
%
Technische Universit{\"a}t Berlin - Machine Learning Group \\
Stra{\ss}e des 17. Juni 135, 10623 Berlin - Germany
}

\maketitle

\begin{abstract}
Change-point detection (CPD) in high-dimensional, large-volume time series is challenging for statistical consistency, scalability, and interpretability. We introduce \textit{TimePred}, a self-supervised framework that reduces multivariate CPD to univariate mean-shift detection by predicting each sample's normalized time index. This enables efficient offline CPD using existing algorithms and supports the integration of XAI attribution methods for feature-level explanations. Our experiments show competitive CPD performance while reducing computational cost by up to two orders of magnitude. In an industrial manufacturing case study, we demonstrate improved detection accuracy and illustrate the practical value of interpretable change-point insights.
\end{abstract}

\section{Introduction}
Change-point detection (CPD) identifies abrupt distributional shifts in time series \cite{basseville1993detection}. Traditional methods often assume moderate sample sizes and low dimensionality, but modern applications generate high-volume, high-dimensional data, creating challenges in both statistical reliability and computational efficiency \cite{truongSelectiveReviewOffline2020}. Beyond detection, actionable insights require interpretability: practitioners need to understand not only when a change occurs, but also why \cite{aminikhanghahiSurveyMethodsTime2017, leeExplainableTimeSeries2024}.

To address these challenges, we propose \textit{TimePred}, a self-supervised framework that transforms high-dimensional CPD into a univariate problem by learning to predict each sample's time index. This allows efficient offline detection using existing algorithms, while integrated eXplainable AI (XAI) \cite{samek_explaining_2021} methods highlight the features driving detected changes. We evaluate our method with a set of quantitative experiments and present an application case from the industrial manufacturing domain. There, offline CPD is particularly useful for segmenting production process data and post-production quality assurance or root cause analysis \cite{xuChangepointDetectionDeep2025}. In addition to the mere detection of defects, our XAI insights complement detailed contextual information which is essential to enable effective and timely decision-making.

\section{Related Work}
\label{sec:related}
We briefly review relevant CPD methods and utilized baselines, following \cite{truongSelectiveReviewOffline2020}. CPD segments a multivariate time series $\{x_t \in \mathbb{R}^d\}_{t=1}^T$ into $K$ intervals by minimizing $V(T,x) = \sum_{k=0}^K c(x_{t_k:t_{k+1}})$, where cost function $c(\cdot)$ measures similarity within segments. Common cost functions include: mean-shift cost ($c_{L_2}$), which sums the squared deviations from the segment mean and efficiently detects shifts in average values; autoregressive cost ($c_{ar}$), which models each point as a linear combination of past values and captures changes in temporal dependencies; and kernel-based cost ($c_{rbf}$), which evaluates pairwise similarity between points to detect complex, nonlinear changes without parametric assumptions. Each cost balances sensitivity to different types of change with computational complexity.


Assuming a known number of CPs, $V(T,x)$ can be minimized via dynamic programming with complexity $\mathcal{O}(C K T^2)$, where $C$ is the complexity the considered cost function. Consequently, mean-shift models are far faster than auto-regressive or kernel methods (the latter scale quadratically with segment length). Furthermore, all three cost functions scale linearly with the number of dimensions. In practice, high-dimensional, high-volume data are therefore often pre-processed and each dimension treated individually \cite{letzgusChangepointDetectionWind2020}. However, this misses more complex changes, such as in variable interactions \cite{Jirak2012ChangePoint}. Recent approaches extend CPD to high dimensions: sparsified binary segmentation \cite{Cho2014MultipleChange} amplifies relevant features; projection-based methods \cite{Wang2018HighDimensional} reduce to univariate detection via singular vectors; wavelet transforms \cite{Barigozzi2018Simultaneous} convert high-dimensional CPD into mean-shift detection. \textit{TimerPred} builds upon these ideas, reducing high-dimensional CPD to univariate mean-shift detection.

\section{Method}

\begin{figure}[h!]
\centering
\includegraphics[scale=0.54]{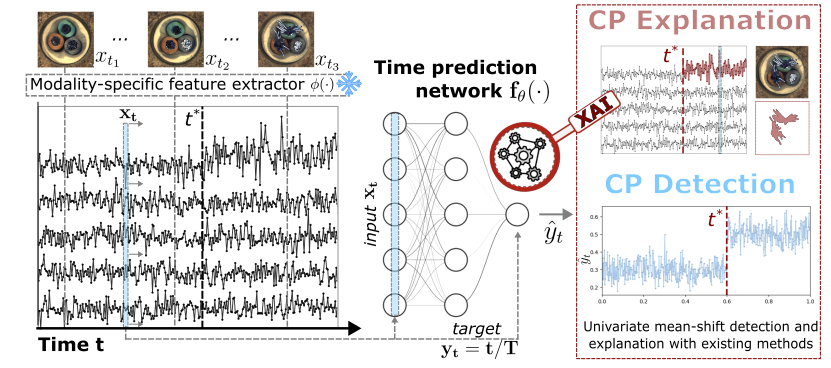}
\caption{Overview on our \textit{TimePred} approach. The time prediction network allows to detect CPs (at $t^*$) on its univariate output and enables XAI integration.}\label{fig:overview}
\end{figure}

\textit{TimePred} utilizes machine learning techniques from self-supervised~\cite{chen2020simclr} and representation learning~\cite{bengio2013representation} to reduce the complexity of CPD tasks. Inspired by early self-supervision methods in computer vision~\cite{doersch2015unsupervised}, which leverage spatial context between image patches, we propose a \emph{temporal self-supervised regression task}: predict the (normalized) time index \(t/T\) of each observation from its feature representation. We train a time prediction model \(f_\theta: \mathbb{R}^d \to \mathbb{R}\) to minimize $
\mathcal{L}(\theta) = \frac{1}{T} \sum_{t=1}^T \ell(f_\theta(x_t),\, \frac{t}{T}),$
where \(\ell(\cdot,\cdot)\) is a regression loss (e.g., mean squared error). To prevent overfitting to spurious temporal noise, we apply strong regularization on \(f_\theta\). For data without structural changes the model is expected to simply fit the average normalized time index ($\mathbb{E}(\frac{t}{T})=0.5$). In the presence of a CP at time $t^*$, however, the model can utilize this information to predict the distinct expected time indices for the respective segment k ($\mathbb{E}_k=\frac{t_{k-1}^*-t_{k}^*}{(K+1)T}$)---yielding observable \emph{mean shifts} in $y_t := f_\theta(x_t)$. 

Once the scalar sequence \(\{y_t\}\) is obtained, existing univariate CPD methods can be directly applied to detect changes in its mean (see Sec.\ref{sec:related}). Note, that the approach works best for offline CPD, as in an online setting, we would need to fit $f_\theta$ after each incoming data point. Furthermore, we can utilize pre-trained backbone models (\(\phi(\cdot)\)) as frozen feature extractors, which allows CPD across data modalities. Beyond simplifying the CPD problem, our framework enables interpretability via eXplainable AI (XAI) methods \cite{samek_explaining_2021}. Specifically, attribution techniques can be used to identify which input features are most responsible for the detected changes. The framework is depicted in Fig.\ref{fig:overview}.

\section{Experiments \& Results}

\begin{figure}[h!]
    \centering
    \includegraphics[scale=0.33]{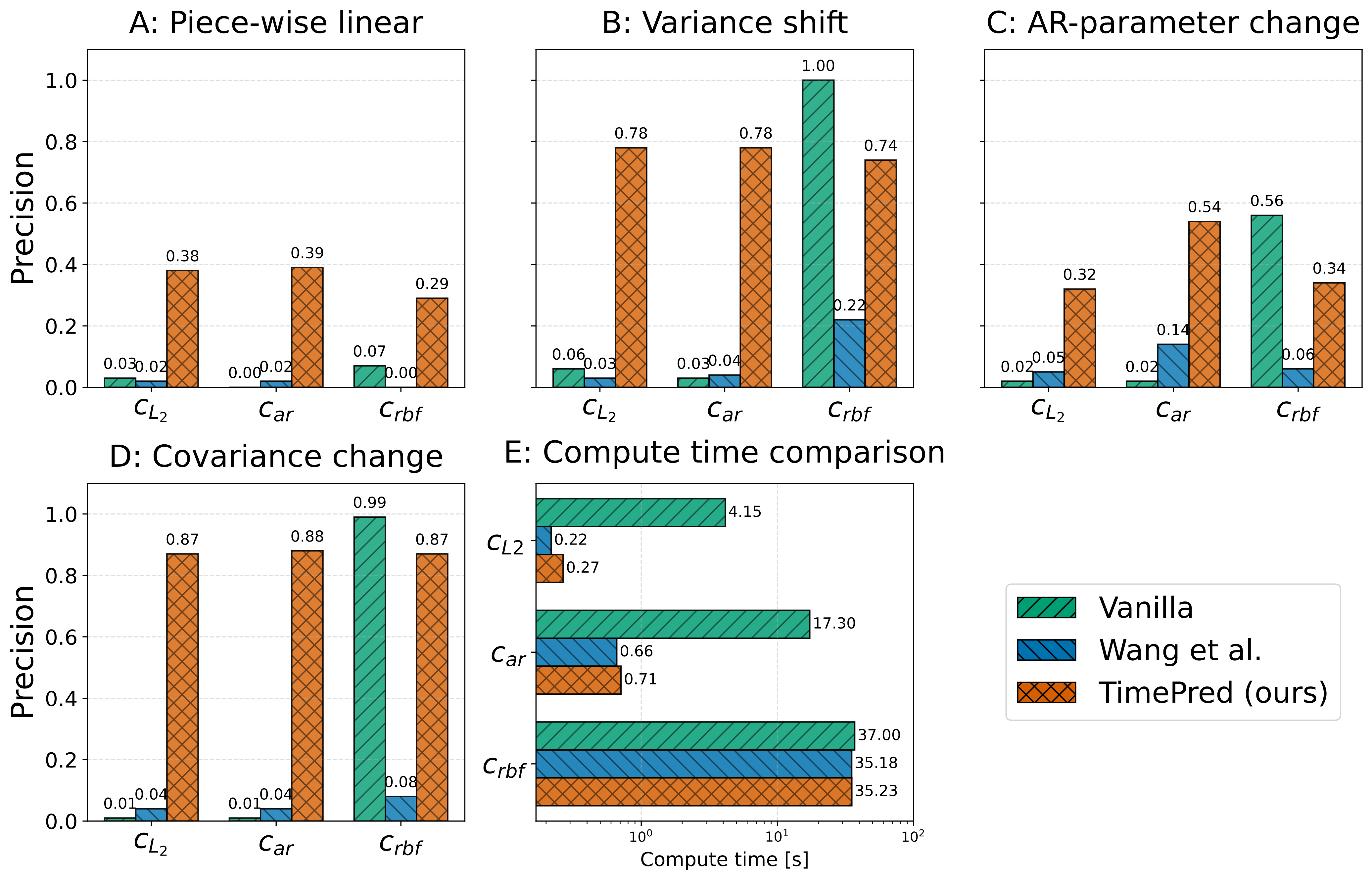}
    \caption{\textbf{A-D}: quantitative results per CP problem (mean precision over 100 data sets each, higher is better). The mean-shift problems were detected perfectly across all methods and are therefore not displayed. \textbf{E}: comparison algorithm runtime in seconds (lower is better).}
    \label{fig:results}
\end{figure}

\subsection{Quantitative Experiments}
We generate a set of multivariate CPD problems, each of shape (T=10.000, d=100), in which a subset of the dimensions exhibit \textit{one} abrupt change (randomly drawn) with one of the following properties:

\textbf{Mean shift:} one of the dimensions exhibits a shift in the mean.

\textbf{PW linear:} one dimension is piecewise linear based on (Gaussian) others.

\textbf{Variance shift:} $\sigma^2$ is resampled for half of the normally distributed signals.

\textbf{AR shift:} autoregressive (AR) coefficients (order=1) change abruptly.

\textbf{Cov shift:} Gaussian signals where the covariance changes abruptly.

We repeat the data generation process 100 times per setting and report the average precision of detection within a tolerance window of $\pm{100}$ steps around the true CP. Furthermore, we compare the runtime of different configurations to evaluate computational efficiency.

We compare CPD on the raw data (\textit{vanilla}), the projected signal following \cite{Wang2018HighDimensional} (\textit{Wang et. al.}), and our proposed predicted time index (\textit{TimePred}) utilizing $c_{L_2}$, $c_{ar}$ and $c_{rbf}$ and dynamic programming as implemented by \cite{truongSelectiveReviewOffline2020}. As time prediction network ($f_\theta$), we utilize a three-layer, L1 and L2 regularized MLP. Experiments were run on an Apple M4 Max processor (16-core CPU, 64 GB RAM) with no hardware-specific tuning. For details, see published code \cite{PredTime2025_repo}.

Results are reported in Fig.\ref{fig:results}. Among the baselines, only the computationally expensive \textit{vanilla} rbf-kernel was able to produce competitive results. It performed overall best in variance, ar-parameter and covariance change detection, closely followed by our \textit{TimePred} algorithm. Our method took the lead for piece-wise linear signals. Notably, \textit{TimePred} achieves its top performance by utilizing the computationally more efficient $c_{L_2}$ and $c_{ar}$ cost functions, which reduces computation time by up to 130x (TimePred $c_{L_2}$ vs Vanilla $c_{rbf})$.

\subsection{Case Study: Image-based Quality Monitoring}
This scenario illustrates automated quality control in industrial manufacturing using \textit{TimePred}. We use the MVTec Anomaly Detection dataset \cite{Bergmann2021MVTecAD}, containing high-resolution images from 15 product categories (e.g., transostor, tile, cable) with and without defects. Images are ordered by anomaly label --- normal first, then defective --- simulating a temporal shift from normal to faulty production. The goal is to detect this CP, indicating the onset of quality degradation. We employ a pre-trained ResNet50 as feature extractor ($g_\beta$) and replace the classifier with an ElasticNet regression head ($f_\theta$). Precision is measured with $\pm$25 time steps of tolerance. For \textit{explaining} the detected CPs we utilize a regression-specific flavor Layer-wise-relevance-propagation (LRP) \cite{bach-plos15, letzgus2022toward}, a performant propagation-based XAI method, to obtain an explicit explanation as to why a sample was assigned to the anomalous part of the series of images. For implementation details see \cite{PredTime2025_repo}.

\begin{figure}[h!]
    \centering
    \includegraphics[scale=0.315]{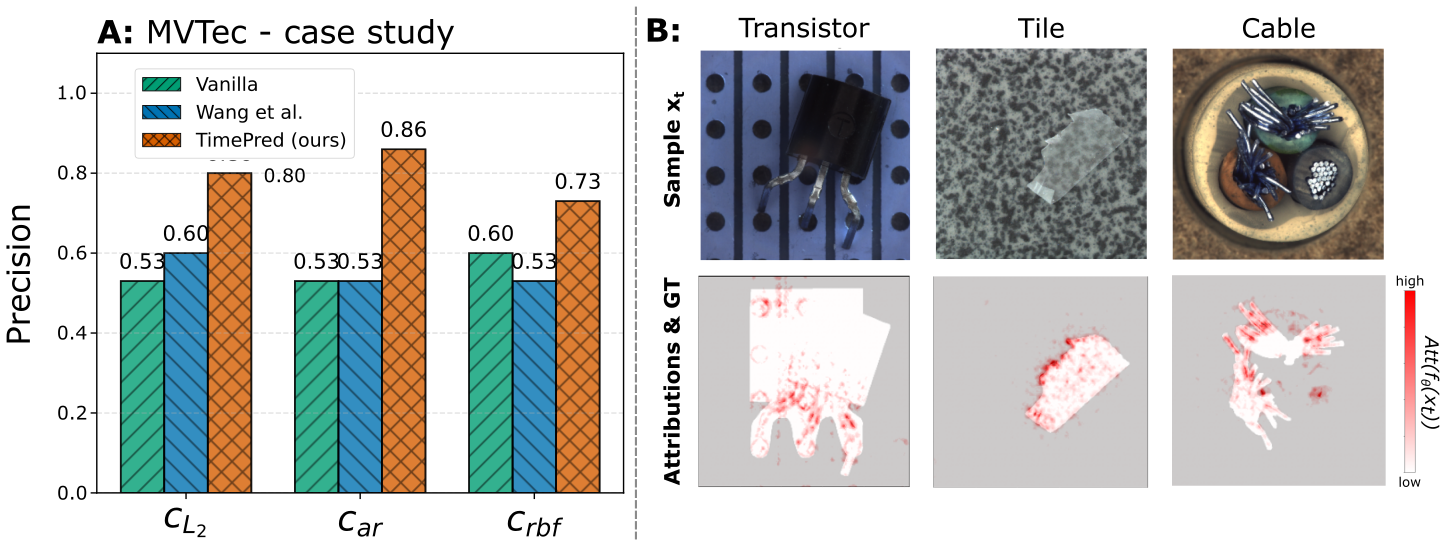}
    \caption{\textbf{A}: CPD results on the MVTec case study. Average precision over 15 objects (higher is better). \textbf{B}: examples of (anomalous) objects from after the detected change point (top) and their corresponding XAI heatmap indicating why the sample was associated with the abnormal segment (bottom). Additionally, the anomaly ground truth (GT) is shown as a grey overlay.}
    \label{fig:MVTec}
\end{figure}

The results are displayed in Fig.\ref{fig:MVTec}. \textit{TimePred} outperforms the closest competitor by more than 40\% in the CPD task. Additionally, we present a selection of LRP heatmaps for the respective objects, alongside the anomaly ground truth masks. We observe high overlap between the two which confirms that the algorithm decided for the change point for the right reasons and gives practitioners actionable insights into \textit{what} has changed.

\section{Discussion \& Outlook}
We have introduced \textit{TimePred}, a framework that enables efficient and accurate CPD in high-volume time series by transforming diverse and complex CPD tasks into tractable univariate mean-shift problems. This reduction yields substantial computational gains and enables seamless integration of XAI methods.

A conceptual limitation of our method is that we cannot expect to detect CPs for a regime that alternates between two states in a periodic manner. The respective univariate time prediction model output would not exhibit any changes. In practice, the performance of \textit{TimePred} is furthermore sensitive to choices in design parameters of the time prediction network and its training. Their influence on the CPD performance should be evaluated more systematically on existing, large-scale benchmarks in the future. 

While we have tested \textit{TimePred} in the setting of detecting one single CP in an offline fashion, future work could focus on extending the method for online CPD and an unknown number of changes. Another promising extension of our work could be to explore the method's natural potential for uncertainty quantification by aggregating CPD results across different time-prediction models.

\begin{footnotesize}

\newpage

\bibliographystyle{IEEEtran}
\bibliography{sample}

\end{footnotesize}


\end{document}